\pdfoutput=1

\documentclass[11pt]{article}

\usepackage{EMNLP2023}
\usepackage{listings}

\usepackage{times}
\usepackage{latexsym}
\usepackage{subcaption}
\usepackage{url}
\usepackage{array}
\usepackage{comment}
\usepackage{subcaption}

\usepackage[T1]{fontenc}

\usepackage[utf8]{inputenc}

\usepackage{microtype}

\usepackage{inconsolata}
\usepackage{graphicx}

\title{GPT Deciphering Fedspeak: Quantifying Dissent Among Hawks and Doves}

\newcommand{\fomc}{\textcolor{black}{\textsc{FOMC}}}
\newcommand{\tr}{\textcolor{violet}{transcripts}}
\newcommand{\tra}{\textcolor{violet}{transcript}}
\newcommand{\mi}{\textcolor{olive}{minutes}}
\newcommand{\st}{\textcolor{teal}{statements}}
\newcommand{\sta}{\textcolor{teal}{statement}}
\newcommand{\doc}{\textcolor{black}{documents}}

\author{
  Denis Peskoff \\ 
  Office of Population Research \\
  Princeton University \\ \texttt{dp2896@princeton.edu}  \\
  \And
 Adam Visokay \\
 Sociology \\  University of Washington  \\
 \texttt{avisokay@uw.edu} \\
\And
 Sander Schulhoff  \\
 Computer Science \\  University of Maryland  \\
\texttt{sschulho@umd.edu} \\
\AND
 Benjamin Wachspress \\
 Computer Science \\  Princeton University  \\
\texttt{bjw6@princeton.edu} \\
  \And
  Alan Blinder \\
  Economics \\
  Princeton University \\
\texttt{blinder@princeton.edu}
  \And
   Brandon M. Stewart \\
  Sociology and OPR \\
  Princeton University \\
  \texttt{bms4@princeton.edu}
}
\begin{document}

\maketitle

\begin{abstract}

Markets and policymakers around the world hang on the consequential monetary policy decisions made by the Federal Open Market Committee (\fomc{}).
Publicly available textual documentation of their meetings provides insight into members' attitudes about the economy. We use GPT-4 to quantify dissent among members on the topic of inflation. We find that \tr{} and \mi{} reflect the diversity of member views about the macroeconomic outlook in a way that is lost or omitted from the public \st{}. In fact, diverging opinions that shed light upon the committee's ``true'' attitudes are almost entirely omitted from the final \st{}. Hence, we argue that forecasting \fomc{} sentiment based solely on \st{} will not sufficiently reflect dissent among the hawks and doves. 

\end{abstract}

\section{The Road to FOMC Transparency}
The Federal Open Market Committee (\fomc{}) is responsible for controlling inflation in the United States, using instruments which dramatically affect the housing and financial markets, among others.
For most of the 20\textsuperscript{th} century, conventional wisdom held that monetary policy is most effective when decision-making was shrouded in secrecy; the tight-lipped Alan Greenspan, a past chairman of the Fed, quipped about \textit{“learning to mumble with great incoherence.”} But times change.

\citet{blinder2008survey} show how the emergence of greater transparency and strategic communication became an important feature of 21\textsuperscript{st} century central banking. 
Fed communication is now an integral component of monetary policy, and ``Fed watchers'' dote on every word.
The \fomc{} first started releasing public \st{} following their meetings in February 1994.
This meager documentation grew and now consists of three types for each official meeting: carefully produced and highly stylized one page \st{} are released immediately after each FOMC meeting, followed about three weeks later by lengthier \mi{}, and finally five years later by full, verbatim \tr{}. Subsequently this triplet is referred to as \doc{}.
We find \mi{} closely reflect the content of \tr{}, so to avoid redundancy, we focus our analysis on \tr{} and \st{}.

\begin{figure}
    \centering
    \begin{subfigure}{.5\linewidth}
     \includegraphics[width=\linewidth]{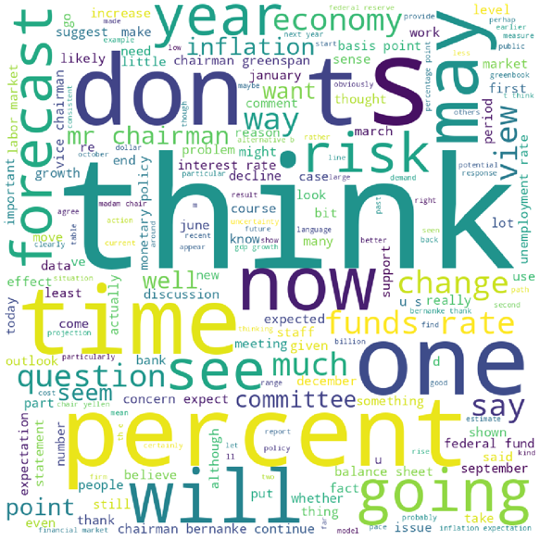}
    \end{subfigure}%
    \begin{subfigure}{.5\linewidth}
        \includegraphics[width=\linewidth]{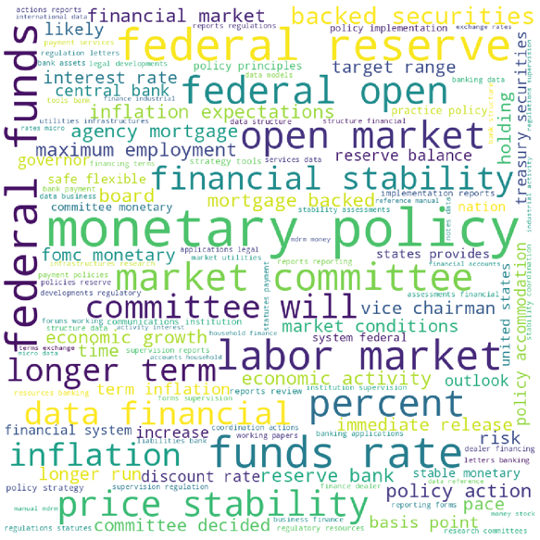}
    \end{subfigure}%
    \centering
    \caption{The \tr{} (left) contains opinions and disagreements but \st{} (right) are concise.  We analyze both datasets using GPT-4 prompting.} %
    \label{fig:words}
\end{figure}

Increased \fomc{} communication has prompted social science research spanning the disciplines of economics, sociology, finance and political science (Section~\ref{sec:litreview}). 
Financial market participants are also keenly interested. 
Billions, if not trillions of dollars are traded on the Fed's words. 
The interpretations---right or wrong---of what the \fomc{} “really means” move markets and affect the economy.
However, relying upon \doc{} as data in the social sciences is a challenge due to the lack of structure and the cost of annotation~\citep{grimmer_stewart_2013,gentzkow2019text,ash_hansen_2023}.

\begin{figure*}
    \centering\includegraphics[width=.9\linewidth,height=9.5cm]{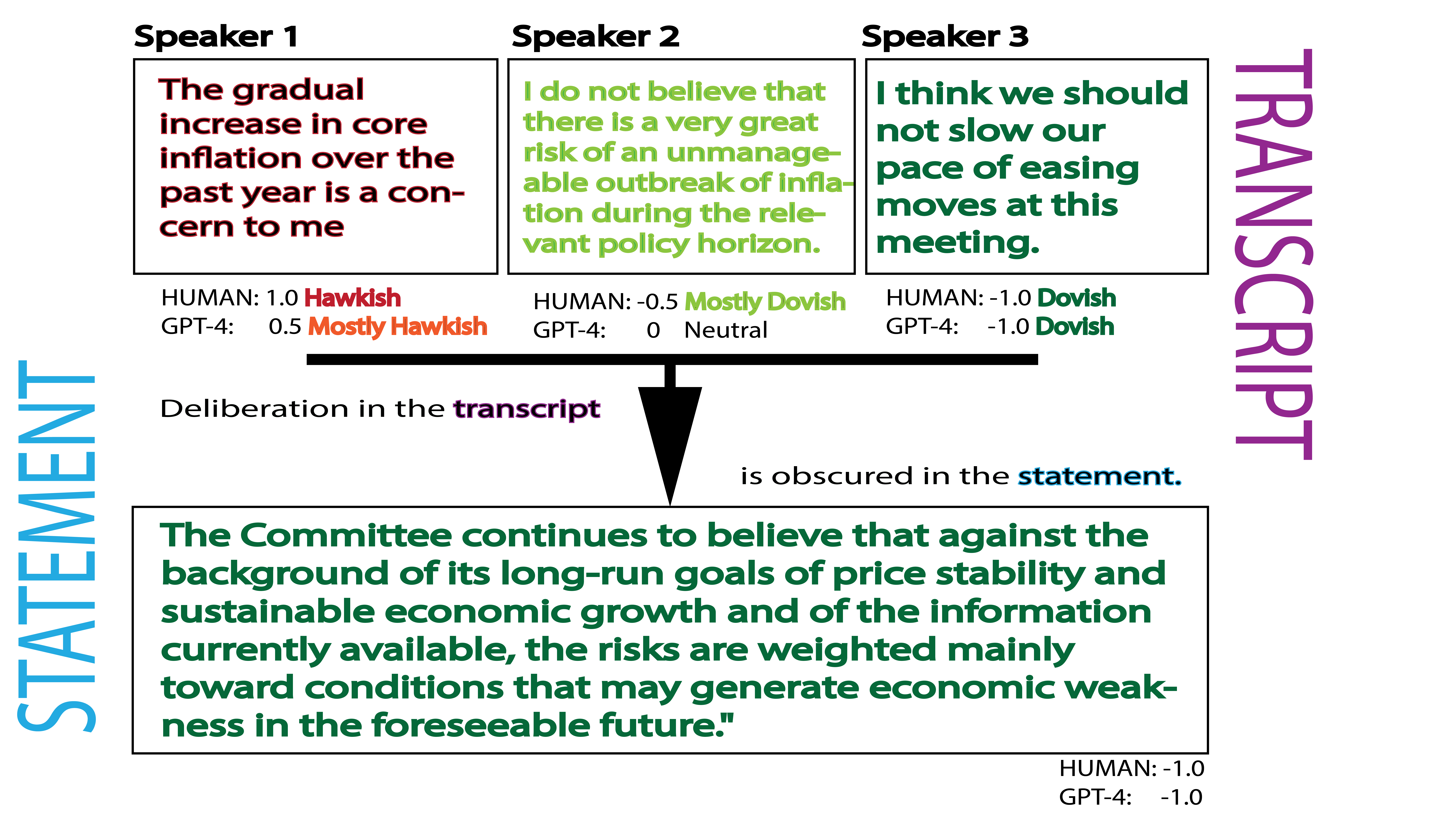}
    \caption{Text taken from an FOMC meeting on December 11, 2001. A dovish \sta{} does not reflect the hawkish sentiment of Speaker 1. GPT-4 can quantify dissent lost from \tr{} to \st{}.}
    \label{fig:figure1}
\end{figure*}

\citet{gpt_decipher_2023} show that Generative Pre-training Transformer (GPT) models outperform a suite of commonly used NLP methods on text quantification.
Motivated by these results, we set out to quantify the language of the \fomc{} using GPT-4 \cite{openai2023gpt4} by preparing %
a combined data set of \fomc{} \doc{} from 1994-2016 (Section~\ref{sec:data}).\footnote{\url{https://github.com/DenisPeskoff/FedNLP}}
We conclude that \tr{} contain more dissent than \st{} (Section~\ref{sec:results}).

\section{\fomc{} Data: Transcript to Statement}
\label{sec:data}

The \fomc{} normally meets eight times per year in order to assess current economic conditions, ultimately deciding upon the path for monetary policy. 
We aggregate and release the official publicly available text documenting these deliberations by the Fed as an aligned corpora of \doc{} from 1994 to 2016.\footnote{\url{https://www.federalreserve.gov/monetarypolicy/fomc_historical.htm}} 
These text \doc{} are similar in content, but \tr{} and \st{} are dramatically different in style and detail (Figure~\ref{fig:words}).\footnote{Transcripts only became being released in 1994 and due to the 5 year lag the latest transcripts are not yet available.}
For our purposes, the \st{} required no pre-processing. For the \tr{}, we use regular expressions to partition and then re-aggregate the text by each unique speaker.
See Appendix~\ref{sec:docs} for an example of each document type.

\subsection{Lessons from Social Science}
\label{sec:litreview}
Past work has used at most one form of \fomc{} meeting documentation, but \textit{rarely multiple in conjunction}. 
For example, in the finance literature, \citet{Mazis2017} apply Latent Semantic Analysis to \fomc{} \st{} to identify the main "themes" used by the committee and how well they explain variation in treasury yields. 
\citet{GU2022216} use \mi{} to investigate how the tonality of committee deliberations impacts subsequent stock market valuations. 
Political scientists use \tr{} to estimate committee members' preferences on inflation and unemployment~\cite{baerg_lowe_2020}. 
Economists have assessed the role of communication in achieving monetary policy objectives by looking at similar documents~\citep{romer2004new,handlan2020text}. 
\citet{HUPPER2023103497} investigate the extent to which shifting inflation focus is reflected in full \tr{}.  
\citet{Hali2021} apply Latent Dirichlet Allocation (LDA) to \tr{} to detect the evolution of prominent topics.
\citet{hansen2017} use LDA to quantify \tr{} and identify how transparency affects the committee's deliberations.

\subsection{Hawks and Doves}
\label{sec:hawkdove}

We take the \tr{} to best represent \fomc{} members' underlying attitudes and think of the \st{} as stylized representations of what they wish to communicate publicly. 
To identify disagreement, we need to go beyond the \st{} and look closely at the language employed by members in their remarks throughout meeting \tr{}.
Dissenting votes are rare because of a strong historical norm: members dissent only if they feel very strongly that the committee’s decision is wrong. 
Modest disagreements do not merit dissent.\footnote{Notably, this is not true of the monetary policy committees of other nations; on some of them, dissent is common—even expected~\citep{tillmann2021financial}.}
That said, members of the committee do frequently voice detectable disagreements with one another at meetings. Disagreements frequently concern the state of the economy, the outlook for inflation, and many other things, including where the range for the federal funds rate should be set that day. 
Such debate and deliberation among members is a routine and productive element of the meetings. 
These disagreements are more clearly expressed in the \tr{}. 
Daniel Tarullo’s comments in a 2016 \tra{} illustrate the point: 
 \begin{quote}
 \textit{“it is institutionally important for us to project an ability to agree, even if only at a fairly high level, and that is why I abstained rather than dissented over each of the past several years [...] I have gone out of my way in the past four years not to highlight publicly my points of difference with the statement.”}
 \end{quote}

\begin{figure*}[t!]
    \centering\includegraphics[width=.95\linewidth]{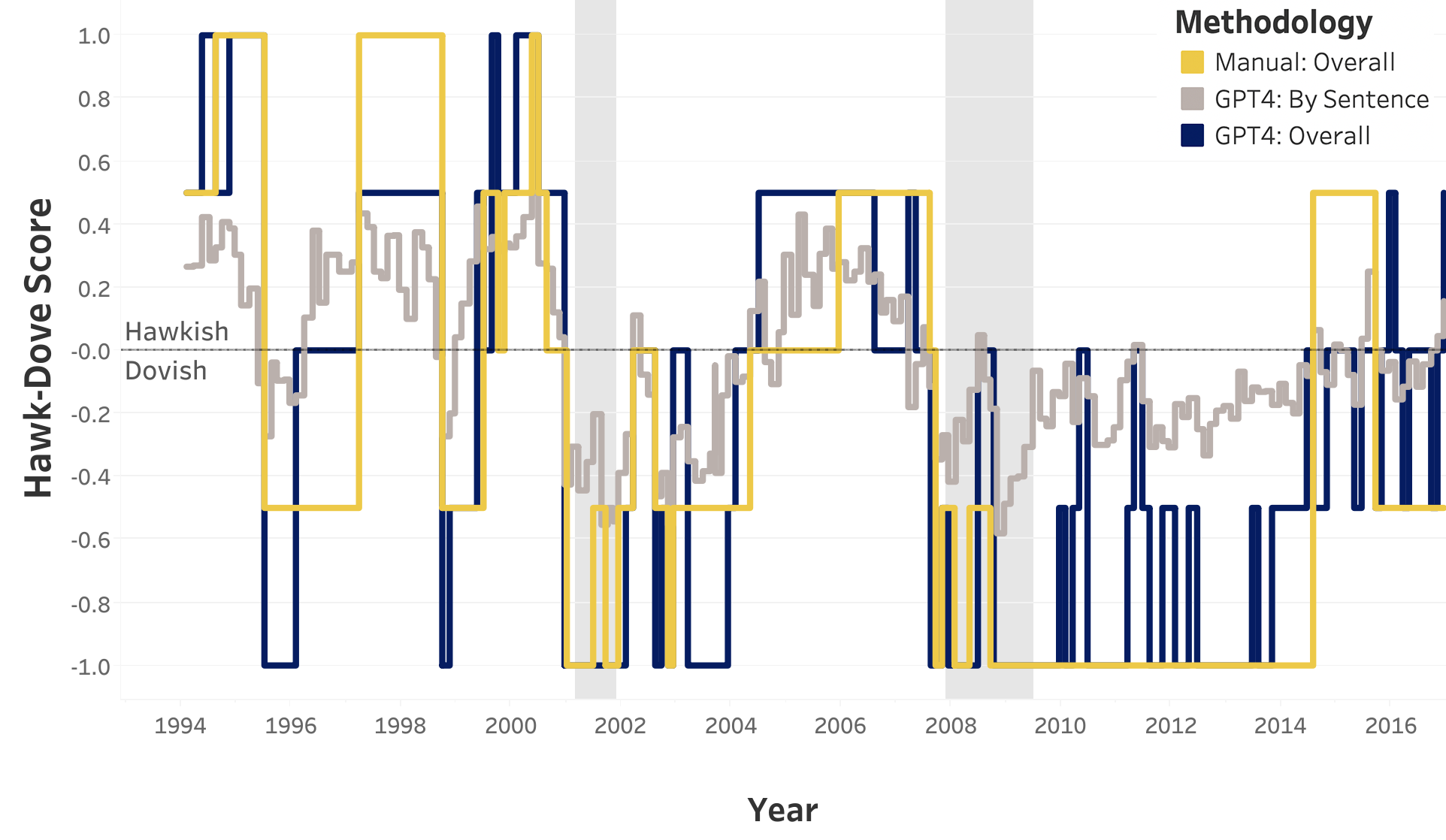}

\caption{Our analysis of \st{} finds that averaging at the sentence-level (gray) loses information since the average sentence is Neutral.  Ingesting the overall statement (blue) better mirrors the manual gold label (gold).}
\label{fig:results}
\end{figure*}

\subsection{Manual Analysis to Create a Gold Label}
\label{sec:methodology}

\begin{figure*}[t!]
    \centering\includegraphics[width=\linewidth]{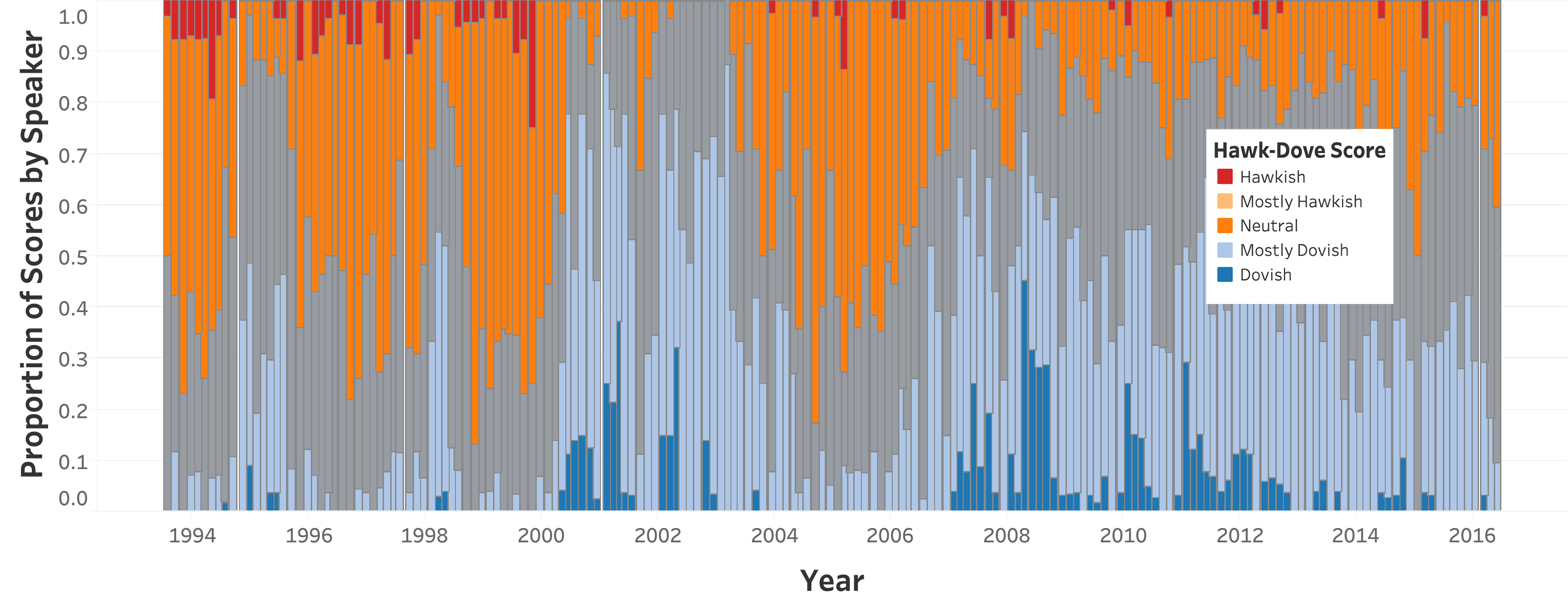}
    \caption{Our analysis of \tr{} with GPT-4 at the speaker level shows that meetings consistently have dissenting opinions.  Similarly to the previous figure, 2001 to 2004 is dovish (blue), while 2004 to 2006 is hawkish (orange).}

\end{figure*}

Dissent amongst speakers is normally concentrated on the discussion of the economic and financial situation of the U.S, specifically inflation targeting. 
For example, in the January 2016 meeting \tra{}, the committee discusses their 2 percent inflation projection in the context of factors such as oil prices, the job market, and the Chinese economy. 
Many of the members argue that the inflation projection of 2 percent will not be accurate, while others who do support the 2 percent projection qualify their support with varying degrees of uncertainty. 
In this meeting, President Mester concludes, 
\begin{quote}
    \textit{“My reasonable confidence that inflation will gradually return to our objective over time recognizes there is and has always been large uncertainty regarding inflation forecasts.”}
\end{quote}
While Mr. Tarullo argues in opposition, \begin{quote}
    \textit{“%
    \dots I didn’t have reasonable confidence that inflation would rise to 2 percent. Nothing since then has increased my confidence. To the contrary, a few more doubts have crept in.”}
\end{quote}
If the meeting \sta{} was an accurate representation of what transpired at the meeting, it would follow that the uncertainty of the committee regarding their inflation forecast would be communicated. 
Instead, the diverging individual opinions are omitted in the final \sta{}, where:
\begin{quote}
\textit{“The Committee currently expects that, with gradual adjustments in the stance of monetary policy, economic activity will expand at a moderate pace and labor market indicators will continue to strengthen. Inflation is expected to remain low in the near term, in part because of the further declines in energy prices, but to rise to 2 percent over the medium term as the transitory effects of declines in energy and import prices dissipate and the labor market strengthens”} %
\end{quote}

We apply this level of granular analysis to create a gold label for \st{}, classifying each one according to Hawk and Dove definitions proposed by \citet{gpt_decipher_2023}: Dovish (-1.0), Mostly Dovish (-0.5), Neutral (0), Mostly Hawkish (.5), Hawkish (1.0).
A trained undergraduate does a first pass and escalates any borderline cases to a \fomc{} expert for adjudication.  
We manually review the January 2016 \tra{} and pair it with a simple computational analysis, which finds that most dissent at that time surrounded topics relating to inflation (Appendix~\ref{appendix:comp}).

\subsection{GPT-4 “Reads” Terse Documents}

GPT-4, and Large Language Models more broadly, are a suitable tool for rapid linguistic processing at scale.
We produce three different measurements of hawk/dove sentiment using \st{} and two using \tr{}. %
\citet{gpt_decipher_2023} use GPT-3 to quantify 500 sentences selected uniformly at random from \fomc{} \st{} between 2010 and 2020. We extend this by using GPT-4 to quantify \textbf{all} 3728 sentences in \st{} from 1994 to 2016 (Appendix~\ref{sec:gpt}). 
A limitation of this approach is that the holistic sentiment of the meeting is not captured because each sentence is scored independently---without context.

The first \sta{} measurement we propose is to simply take an unweighted mean of all individual sentence scores for each meeting. 
Because most sentences have nothing to do with inflation (62\% of sentences scored as Neutral), we hypothesize that this method has further limitations. 
We resolve this by ingesting each \textit{entire} \sta{} into our GPT-4 prompt. Both of these measurements, along with manual gold labels, can be seen in Figure~\ref{fig:results}. 

For the final \sta{} measurement, we construct a logit-scaled score \citep{lowe2011},

\[\theta^{(L)} = \log \left( \frac{Hawk + 0.5}{Dove + 0.5} \right) \]

where \textit{Hawk} and \textit{Dove} are the sums of the hawkish and dovish scores, respectively. 
In this approach $\theta^{(L)}$ ignores sentences scored as Neutral, placing more emphasis on the \textit{relative} rather than \textit{absolute} differences between hawkish and dovish sentiment.
Furthermore, since $\theta^{(L)}$ has no predefined end points, this allows us to generate positions at any level of extremity, which more appropriately reflects the outlier meetings. 

When measuring hawk/dove sentiment using \tr{}, the vast amount of text adds an additional challenge.
Rather than evaluating a 100+ page \tra{} at the sentence level, we instead evaluate \tr{} at the speaker level. That is to say, we use GPT-4 to quantify hawk/dove sentiment for each distinct speaker within each transcript, and then aggregate those by meeting date.
In contrast to the 3728 individual sentence observations across all \st{}, we evaluate 5691 speaker observations across all \tr{}. 
We then follow the same steps outlined above to calculate unweighted average and logit-scaled scores. %

The logit-scaled measurements for \tr{} and \st{} track one another quite closely over most of the years, demonstrating that GPT-4 is effective at identifying similar content across document types of dramatically different length and style. 
It is worth noting, however, that the logit-scaled \tra{} scores have larger extremes than \st{}, especially the upper bound: (-3.66, 3.37) and (-3.40, 2.40), respectively. 
This underscores how the \fomc{} devotes considerable attention to curating their communication strategy to convey confidence and unity despite dissent among the hawks and doves.
The logit-scaled scores begin to diverge in 2012, with \tr{} trending more neutral/hawkish while \st{} remain mostly dovish.
Hence, we propose a direct comparison.

\section{LLMs Quantify Economic Text}
\label{sec:results}

\subsection{Measuring Dissent}

We can use the sentence-level \sta{} and speaker-level \tra{} scores from GPT-4 to compute a measure of dissent for each meeting using the following algorithm: 
\vspace{-2mm}
\begin{enumerate}
  \setlength{\itemsep}{0pt}
   \setlength{\parskip}{0pt}
    \item From the list of scores for each meeting, count the number of hawkish/mostly hawkish and the number of dovish/mostly dovish scores. 
    
    \item If there is at least one hawkish score and at least one dovish score within the same meeting, assign Dissent = 1. Else, Dissent = 0. 
\end{enumerate}
\vspace{-2mm}

We find that \textbf{47\%} of \st{} and \textbf{82\%} of \tr{} contain dissent. We also compute the conditional probability of a \tra{}{} containing dissent given the associated \sta{} binary, $P(T=1|S=1)$ and $P(T=1|S=0)$. We find that when a \sta{} contains dissent, $P(T=1|S=1)$, the \tra{} agrees more than \textbf{97\%} of the time. However, for \st{} scored as having no dissent, $P(T=1|S=0)$ we find that more than \textbf{69\% }of associated \tr{} are scored as containing dissent. This means that for the 53\% of \st{} that don't show signs of dissenting opinions, there is likely dissent in the \tra{} as evidenced by the speaker-level hawk/dove scores.

\subsection{Conclusion and Next Steps}
Our method of ingesting the entire \sta{} for an aggregate prediction better captures the extremes, which more closely mirror the gold label human annotation and suggests that Large Language Models can avoid the noise in this nuanced context.  The $F_{1}$ score for this comparison is 0.57. While this is rather low as a measure of model ``fit'', it is important to note that the results rarely flip sentiment (from hawkish to dovish, or mostly hawkish to mostly dovish), rather, it just seems to mostly disagree on adjacent categories. See Figure~\ref{fig:results} for a visual comparison of the sentence-level, entire text, and manual scores.
Of note, the inconsistent provision of \st{} and relatively high volatility in hawk/dove sentiment before 2000 is consistent with \citet{meade2008} and \citet{hansen2017} who have also studied the 1993 change in \fomc{} communication strategy.
We demonstrate that GPT-4 can identify the extremes in dissenting hawk and dove perspectives despite the indications of a clear consensus in the \st{}. This empirical finding supports our manual analysis.

While we focus on \tr{} and \st{}, future work may consider an even more fine grained analysis, incorporating \mi{} as well. We found the content of the \mi{} to more closely resemble the \tr{} than the \st{}, but differences do exist and remain underexplored. 

Additionally, we note that GPT-4 scores made more neutral predictions than the gold standard manual labels. To improve upon this, we created a balanced few-shot example using sentences from \fomc{} \st{} not included in our sample --- meetings since 2020. This marginally improved the prediction ``fit'' ($F_{1}$ of 57\% to 58\%), but we expect that this could be improved much further with additional prompt engineering. 

GPT-4 is able to quickly quantify stylized economic text. 
Our results from quantifying dissent support the hypothesis that dissenting opinions on the topic of inflation omitted from \fomc{} \st{} can be found in the associated \tr{}. 
As LLMs continue to improve, we expect that it will be possible to study even more nuanced questions than the ones we answer here.

\section*{Limitations}
Substantively, strategic signaling in the \fomc{} is a challenging topic and this is only an initial investigation. Dissent does not have clear ground truth labels and thus we are reliant on human judgment and our team's substantive expertise on monetary policy. Finally, as with much current research, our work relies on OpenAI's GPT API, which poses challenges to computational reproducibility, as it relies on the stability of an external system that we cannot control.

\section*{Ethics Statement}
The \fomc{} is a high-stakes body whose activities are already subject to substantial scrutiny. There is some ethical risk in exploring linguistic signals of hidden information. For example, based on the substantive literature we believe that dissent is intentionally signalled in meetings in order to set up future discussions or lay claims to particular positions. Nevertheless, attribution of intention (as implied by 'dissent') always involves some level of error that could be uncomfortable to meeting participants who feel mischaracterized. We also emphasize that our approach to capturing dissent would not be appropriate to use outside this specific context without careful validation. Finally, by making the \fomc{} data more easily available to the NLP community, we also assume some ethical responsibility for the potential uses of that data \citep[see e.g.][]{peng2021mitigating}. We spend under \$1500 on computation and under \$1000 on annotation and believe our results to be reasonably reproducible.  We feel that these concerns are ultimately minor given that all participants are public officials who knew their transcripts would ultimately be released. 

\section*{Acknowledgements}

 This material is based upon work supported by the National Science Foundation under Grant \#2127309 to the Computing Research Association for the CIFellows 2021 Project.  We thank the Initiative for Data-Driven Social Science at Princeton University for grant support and OpenAI for technical support.  Visokay is supported by the National Institute of Mental Health of the  NIH  under  Award  Number \#DP2MH122405 and by the Center for Statistics and the Social Sciences at the University of Washington.

\bibliography{anthology,custom}
\bibliographystyle{acl_natbib}

\newpage

\appendix
\section{Appendix}
\label{sec:appendix}

Our appendix contains the GPT prompt we used, a section on our computational analysis, and an example of the statement (Figure~\ref{fig:st}),
transcript (Figure~\ref{fig:tr}), and minutes (Figure~\ref{fig:mi}) for the January 29-30, 2008 meetings.

\section{GPT-4}
\label{sec:gpt}

We used GPT-4 heavily for our experiments and analysis. Recent work \cite{openai2023gpt4, liu2023summary, guan2023cohortgpt} that successfully uses GPT-4 for classification gave us confidence in its quality. Additionally, a human expert on our team examined GPT-4 generated labels, and found that in a sample of 25, our expert agreed with 19 labels with high confidence, and 22 labels with at least moderate confidence.

\subsection{Methodology}

We used variants of a single prompt template for all of our tasks. It contains the relevant labels (Hawkish, Dovish, etc.) as well as their definitions. It includes space for some INPUT and asks which label best applies to the input. When processing different \doc{}, such as \tra{}s, we would switch out the word \st{} in the prompt for the appropriate document word.

For \st{}, we 0-shot prompted GPT-4 to label each statement as one of the five labels. We also ran an experiment in which we classified each sentence of each \st{} as one of the labels, then averaged the sentence scores to get to statement score. Finally, we reran this sentence classification with a 10-shot prompt. The prompt was similar to below, except with 10 examples of sentences and their classifications at the beginning.

For \mi{}, we 0-shot prompted GPT-4-32K to label each statement as one of the five labels.

For \tr{}, we 0-shot prompted GPT-4-32K to examine all of each speakers speech, and provide each speaker a single label for each transcript.

For all API calls to OpenAI, we only modified the model (either GPT-4 or GPT-4-32K). We did not change any other settings.

\begin{lstlisting}[breaklines]

prompt_template = """
<statement>
INPUT
</statement>
<labels>
Dovish: Strongly expresses a belief that the economy may be
growing too slowly and may need stimulus through mon-
etary policy.
Mostly dovish: Overall message expresses a belief that the economy may
be growing too slowly and may need stimulus through
monetary policy.
Neutral: Expresses neither a hawkish nor dovish view and is
mostly objective.
Mostly hawkish: Overall message expresses a belief that the economy is
growing too quickly and may need to be slowed down
through monetary policy.
Hawkish: Strongly expresses a belief that the economy is growing
too quickly and may need to be slowed down through monetary policy.
</labels>
Which label best applies applies to the statement (Dovish, Mostly Dovish, Neutral, Mostly Hawkish, Hawkish)?
"""
\end{lstlisting}

\section{Computational Analysis}
\label{appendix:comp}
\begin{figure*}[t!]
    \centering
    \includegraphics[width=\linewidth]{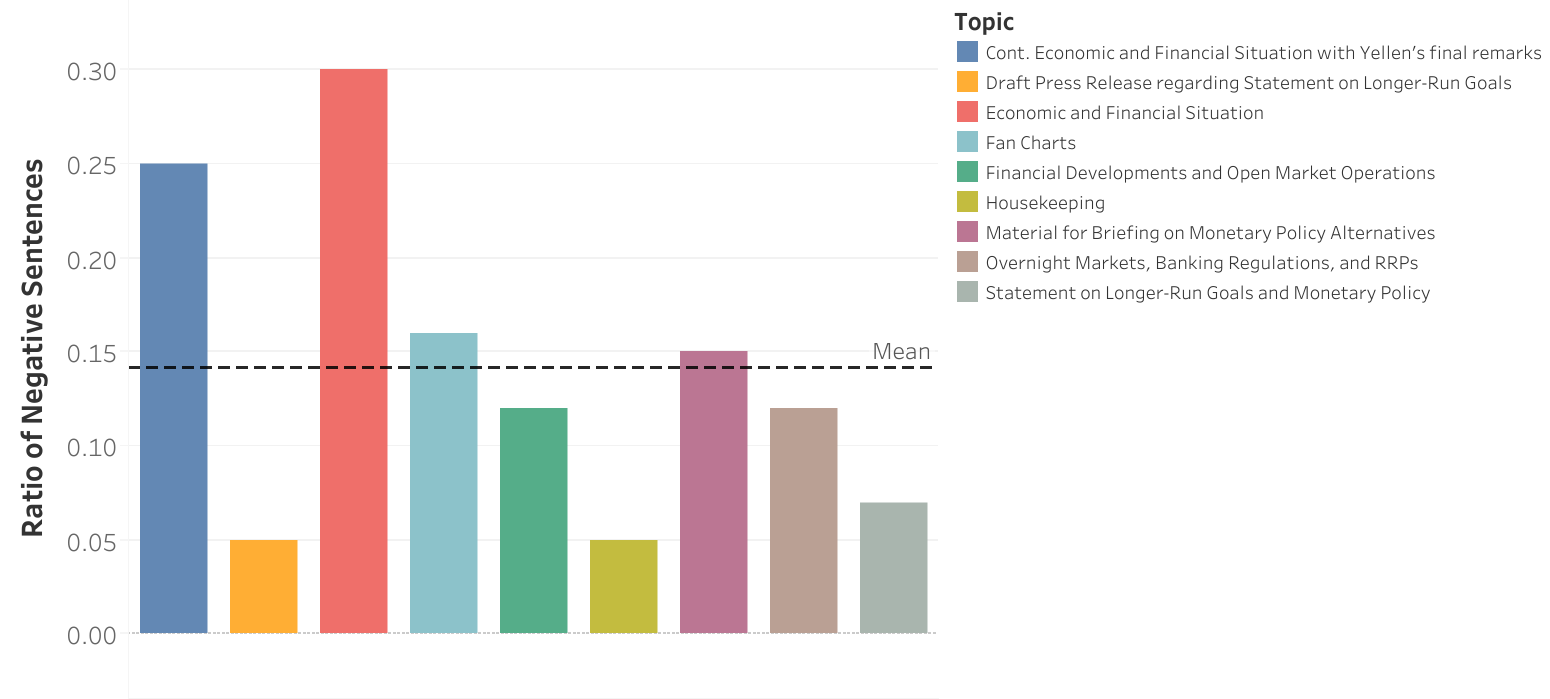}
    \caption{Discussion of inflation (Red and Blue topics) is more contentious than other topics and the average of all data (dotted line).}
    \label{fig:plot}
\end{figure*}

We paired our manual review of the January 26-27, 2016 \tra{} with a computational analysis of dissent in the meeting. We stratified the meeting into nine topics, each corresponding to a portion of the \tra{} content. As a baseline, we counted the number of speakers in each section to see if this metric could reflect dissent. This technique, however, seemed to reflect the length of the conversation as opposed to the degree to which members disagreed with one another. 

Our next approach was to do a sentiment analysis of each topic to see if the prevalence of negativity could indicate dissent. We supposed that negative sentiment would be high if the speakers opposed the stance of either other individuals or the committee as a whole. Using the VADER lexicon~\citep{Hutto_Gilbert_2014}, we calculated the sentiment of each sentence within the nine topics. Since VADER is trained on web-based social media content, which is typically more abrupt than the formal language appearing in the \fomc{} \tra{}, we conducted the sentiment analysis by sentence to optimize the method’s performance. 

To analyze dissent more specifically, we computed the fraction of negative sentences in each topic. For this analysis, we set the threshold negativity score to be 0.1. That is, sentences with a negativity score of 0.1 or higher were classified as negative while all others were not. This number determined by manually reviewing what sentences were captured by varying thresholds and evaluating whether or not they conveyed dissent. When the threshold was set too low (0.05), four out of ten randomly selected sentences conveyed dissent. When set too high (0.15), seven out of ten randomly selected sentences conveyed dissent, but many sentences that indicated dissent were omitted. At the threshold of 0.1, still seven out of ten randomly selected sentences conveyed dissent, and more sentences that conveyed dissent were captured.

\section{Document Examples}

\begin{figure*}
\includegraphics[width=\linewidth]{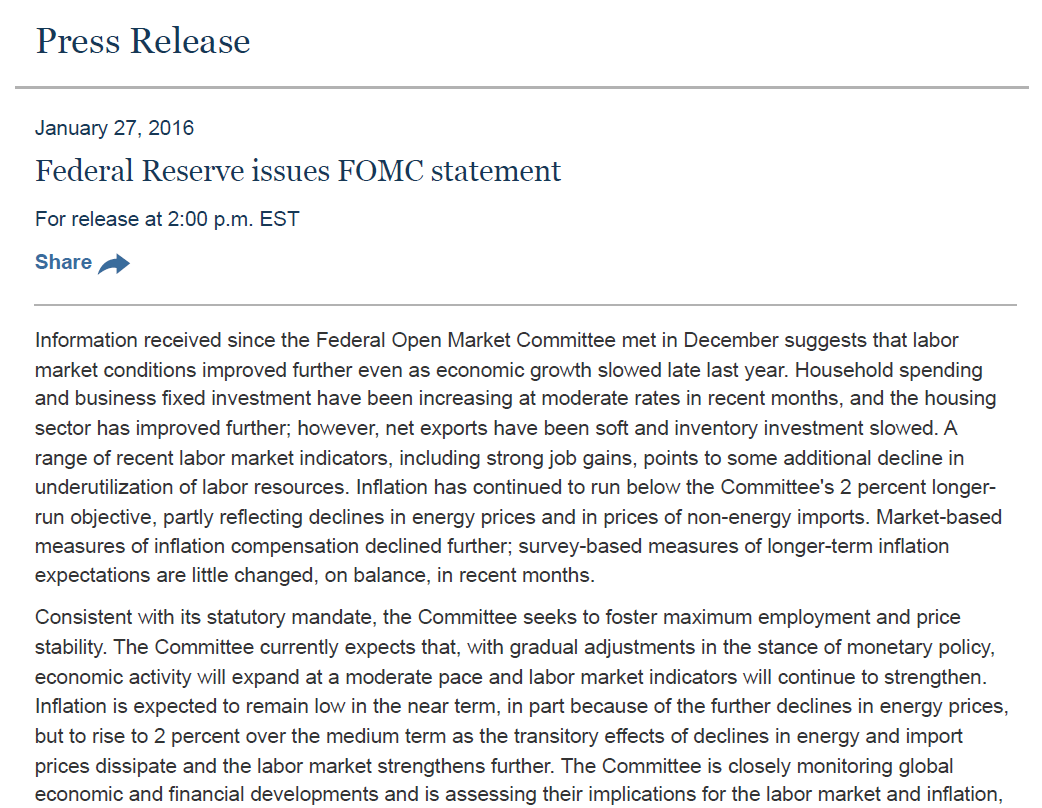}
   \caption{Released statement for the January 26-27, 2016 meeting.}
    \label{fig:st}
\end{figure*}

\begin{figure*}
    \centering
\includegraphics[width=\linewidth]{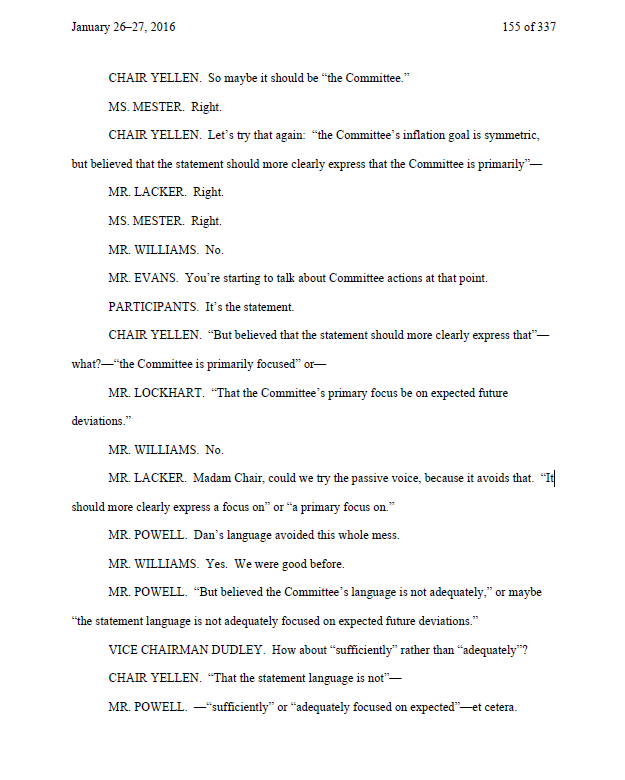}

    \caption{A page of the January 26-27, 2016 meeting transcript that shows dissent.}
    \label{fig:tr}
\end{figure*}

\begin{figure*}
\includegraphics[width=\linewidth]{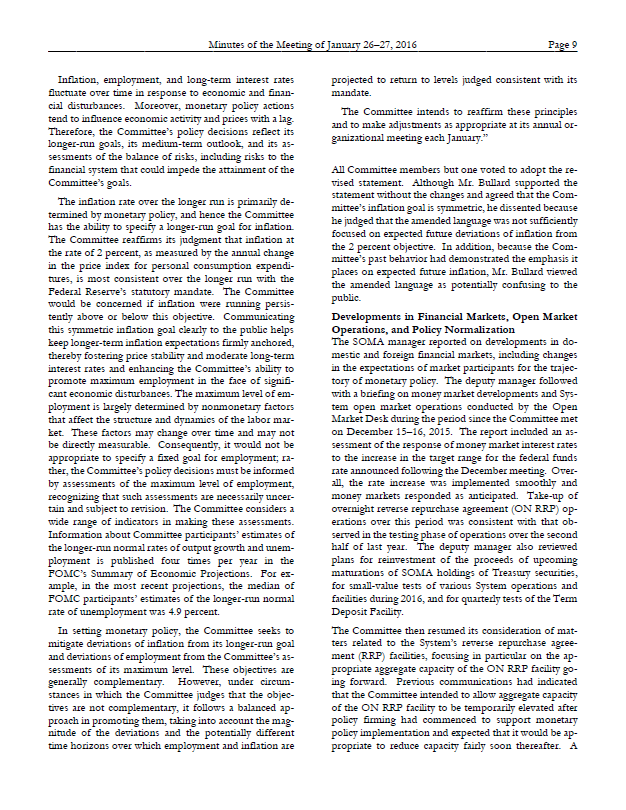}
   \caption{A page of the January 26-27, 2016 meeting minutes that discusses inflation targeting.}
    \label{fig:mi}
\end{figure*}

See Figures~\ref{fig:st}, \ref{fig:tr}, and \ref{fig:mi} for examples of the documents.

\label{sec:docs}

\end{document}